# Towards Classification of Web ontologies using the Horizontal and Vertical Segmentation

Redouane Nejjahi [1], Noreddine Gherabi [2] and Abderrahim Marzouk [1]

[1] Hassan 1st University, FSTS, IR2M Laboratory, Settat, Morocco
[2] Hassan 1st University, ENSAK, LIPOSI Laboratory, Khouribga, Morocco
{nejjahi, gherabi}@gmail.com
amarzouk2004@yahoo.fr

**Abstract.** The new era of the Web is known as the semantic Web or the Web of data. The semantic Web depends on ontologies that are seen as one of its pillars. The bigger these ontologies, the greater their exploitation. However, when these ontologies become too big other problems may appear, such as the complexity to charge big files in memory, the time it needs to download such files and especially the time it needs to make reasoning on them. We discuss in this paper approaches for segmenting such big Web ontologies as well as its usefulness. The segmentation method extracts from an existing ontology a segment that represents a layer or a generation in the existing ontology; i.e. a horizontally extraction. The extracted segment should be itself an ontology.

## 1 Introduction

The Web ontologies present several interests for the Web, such as annotating data [1] [2], distinguishing between homonyms and polysemy, generalizing or specializing concepts, driving intelligent user interfaces and even inferring entirely new (implicit) information [3] [4]. Ontologies are created by ontology engineers with the help of domain experts [5]. Let us take as an example an ontology representing a population. This ontology should have information about the citizens, their dates of birth, relationships, hobbies, addresses, competences, jobs, etc. It seems to be a great ontology allowing us to get new information about one person's tendencies, how these tendencies may be affected by his relationships; also, the companies can use it to target their advertisements. But, for one reason or another, we may not be concerned by the population under a certain age, or may be interested only in a particular city's population. For such purposes, we assume that segments of such big ontologies that contain only the desired information will respond better to the users' expectations.

Several studies are focused on the extraction, classification [6] [7] and segmentation [3] [8] of data in Web ontology; these data can be represented in the ontology web. For example, N. Gherabi et al [9] present a new approach to mapping data stored in relational databases in the semantic Web, it uses simple mappings based on certain specifications of the database schema and explain how relational databases can be used to define a mapping mechanism between the relational database and the OWL ontology. In another work [10], the authors have developed a method to convert UML schemas to Web Ontology.

J. Seidenberg and A. Rector have presented in [3] a method for extracting small segments from large ontologies using the GALEN as an example. The segmentation algorithm they have presented was based on the classes hierarchy, which leads to segmenting the ontology by extracting a specific class hierarchy. Such method responds to segmenting ontologies like the GALEN ontology where one could be interested in a concept like HEART and all its super classes and subclasses.

A. Simonyi and M. Szőts have showed in [11] an ontology segmentation tool which extracts subontologies from a given ontology based on the criterion of relevance supplied by the end user, mainly concept relevance.

In this paper, we aim to present different methods for segmenting OWL ontologies: horizontally (based on criteria on individuals), vertically (based on the choice of classes or properties to maintain) and both.

## 2    Advantages of Web Ontologies Segmentation

The Web ontologies after being created, they grow exponentially in size, either by importing new information or by inferring additional information using the reasoners.

Once these ontologies are sufficiently massive, we will be faced with problems related to their treatment. For example, the Gene ontology is of size 148 MB, the Foundational Model of Anatomy ontology is of size 191 MB, and the Thesaurus ontology is of size 257 MB. Processing such files require machines with a certain level of performance.

By reducing the size of the ontology file, we can reduce the latency of the data load in memory. In addition, files with smaller sizes are highly desirable for the transfer on the Internet or to be processed by machines with limited resources.

Ontology segmentation algorithms aim to reduce the size of the ontology file, some of these algorithms can reduce the size of the ontology file to a quarter the original size which can accelerate the reasoning, especially when the original ontology is a big one like the GALEN ontology which contains 23139 classes [3] [7] [12] [13].

## 3    The horizontal Segmentation Algorithm

The OWL ontologies consist of several parts. Among the most important, we have classes, properties and individuals. By analogy to the relations from relational algebra, we can say that the classes and properties represent the schema of the ontology while individuals represent its extension. Large part of the ontology will be that describing individuals. The segmentation process in the horizontal segmentation will be applied on the individuals to extract those fulfilling the condition(s) of filtering. These conditions may be expressed as range of ObjectProperty = X, or range of DatatypeProperty > Y.

### 3.1    The example ontology

The segmentation algorithm is based on the properties of the ontology to segment.

For example, from an ontology that treats citizens, and therefore defines the following properties: hasAsFather, hasAsMother, hasABondOfBrotherhood, isMarriedTo, livesIn, dateOfBirth, isLoctedIn (city X isLocatedIn country Y), isFriendOf, etc. we can extract other ontologies which are segments of the large ontology, and based on one or more of these properties.

The ontology that we will take as an example is an ontology that describes some citizens, their hobbies, their studies, their political orientations and their parental relationships as well as their dwelling places. Figure 1 illustrates its structure.

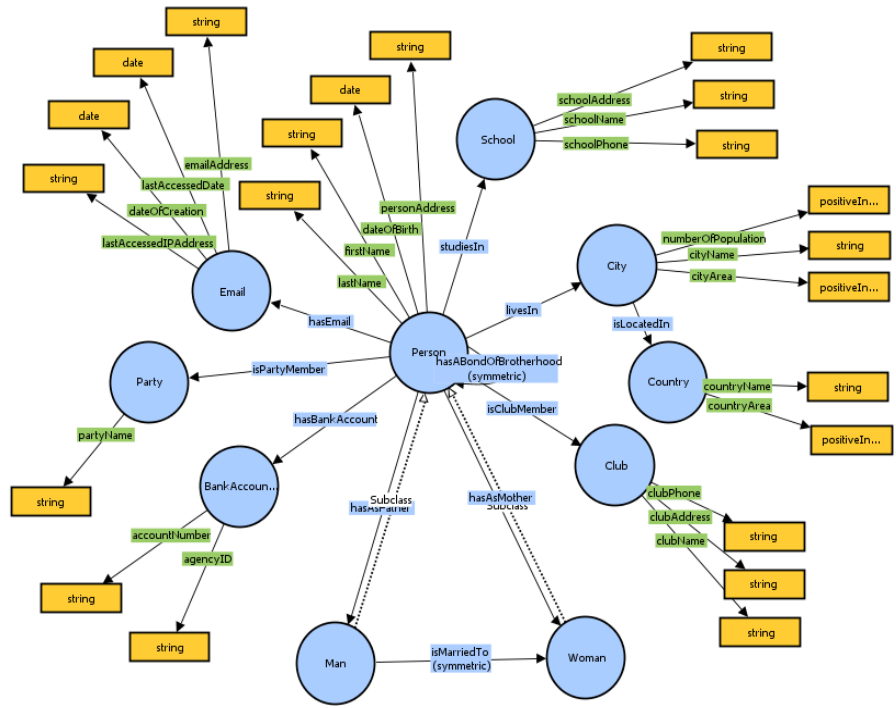

**Fig. 1.** Diagram of the Citizen Ontology.

For the Class and Object Property hierarchies, they are as follows:

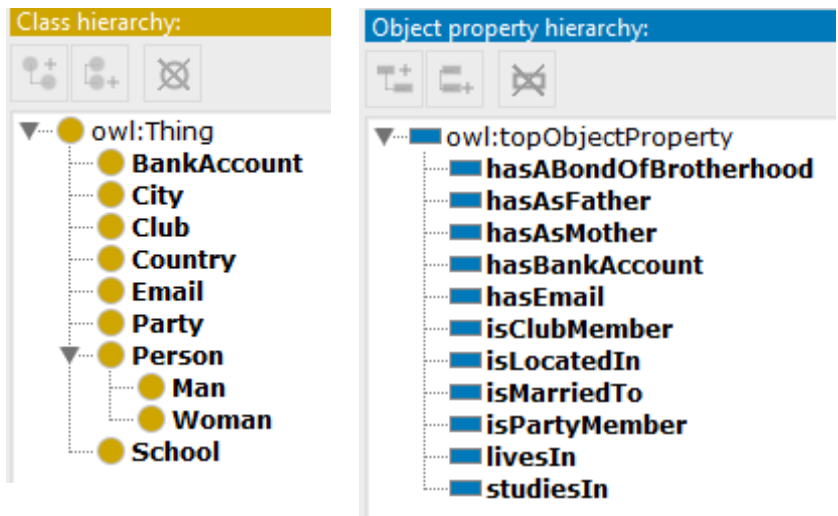

**Fig. 2.** Class and Object property hierarchies.

### 3.2 The ontology Segmentation Algorithm

We start the algorithm with an initial ontology as a data to process. Another input should be the criteria of the segmentation. It can be a condition on a DatatypeProperty e.g. dateOfBirth >= '01/01/1997' or a condition on an ObjectProperty e.g. "live in" -> range = 'Casablanca'. Some of these criteria may be applied directly on the ontology, others may need special processing, e.g. looking for people from Morocco, which leads to look for people with "live in" -> range = x and x "is in" -> range = 'Morocco'.

The algorithm is presented as follows:

```
1 -  Given: SourceOntology, DestinationOntology
2 -  // DestOntology is an empty ontology
3 -  Begin
4 -  CopyStructure (SrcOntology, DestOntology)
5 -  For Each individual In SrcOntology.Individuals
6 -      If (condition = True) Then
7 -          addTo (DestOntology, individual)
8 -      End If
9 -  End For
10 - DeleteAssertionsReferringToNonexistentIndividuals
11 – End
```

We start the algorithm with two files, the first one is the Source Ontology, which represents the whole ontology that we aim to segment, and the second one, called DestinationOntology, is an empty file where we will put the extracted segment.

For our case, the ontology diagram (classes + individuals) contained in the SourceOntology file is shown in Figure 3.

The first step in the algorithm is the CopyStructure procedure, which copies the structure of the source ontology into the destination ontology file. By the structure of the ontology we mean the namespaces declaration, the ontology headers and the classes and properties definitions. This structure, in horizontal segmentation, will be the same for the source ontology and its sub-ontologies.

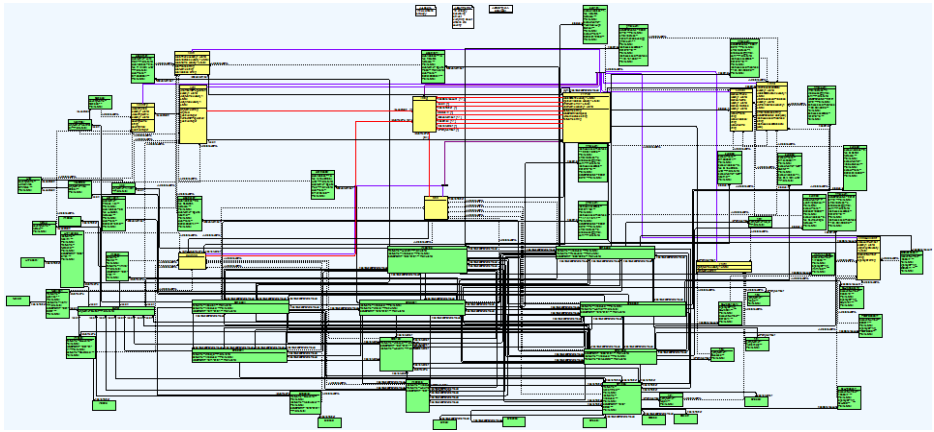

**Fig. 3.** Original Ontology Diagram (Classes + Individuals)

The next step of the algorithm is the main one. It consists of a loop that, for each individual in SrcOntology.Individuals list, checks whether the individual fulfills the condition of filtering or not. This condition may be a combination of conditions connected by logical operators.

Then, if the current individual fulfills the condition, we add it as an individual to the sub-ontology with the addTo procedure. At the end of this step, the ontology diagram (classes + individuals) of the sub-ontology is shown in Figure 4.

The objective of the last stage of the algorithm is the purification of the sub-ontology.

The DeleteAssertionsReferringToNonexistentIndividuals procedure will look for assertions which refer to individuals that have not been selected by the segmentation algorithm. For example, let us say that the criterion of our segmentation algorithm is dateOfBirth < '01/01/1975'. An individual X of dateOfBirth property equal to '16/08/1970' will not show in the resulting ontology; if we had an individual Y where dateOfBirth property is equal to '03/03/2000', it will do. The individual Y may have a relationship with the individual X, so we will end up with a resulting ontology that has an individual referring to a nonexistent one. After adding all the individuals responding to the criterion, the algorithm will filter their Object Properties to maintain only those referring to individuals existing in the resulting ontology to eliminate undesirable information. By the end, our sub-ontology will be as shown in Figure 5

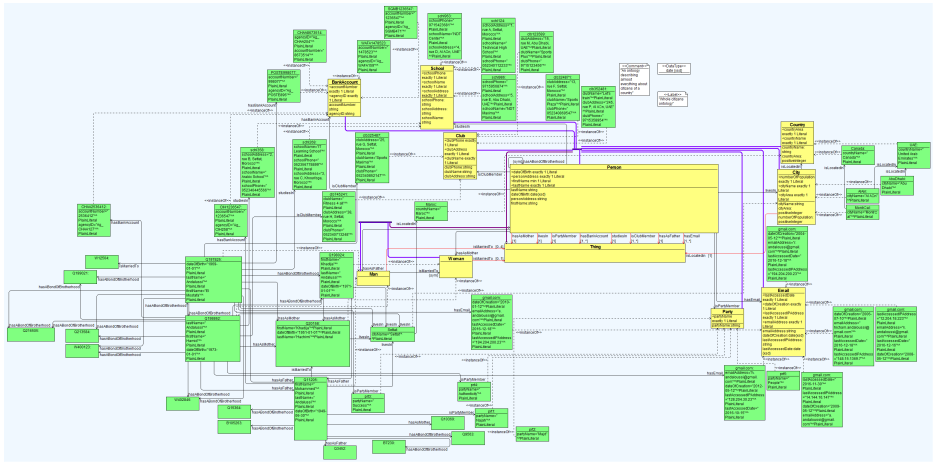

**Fig. 4.** Resultant Sub-Ontology Diagram (Classes + Individuals) before purification

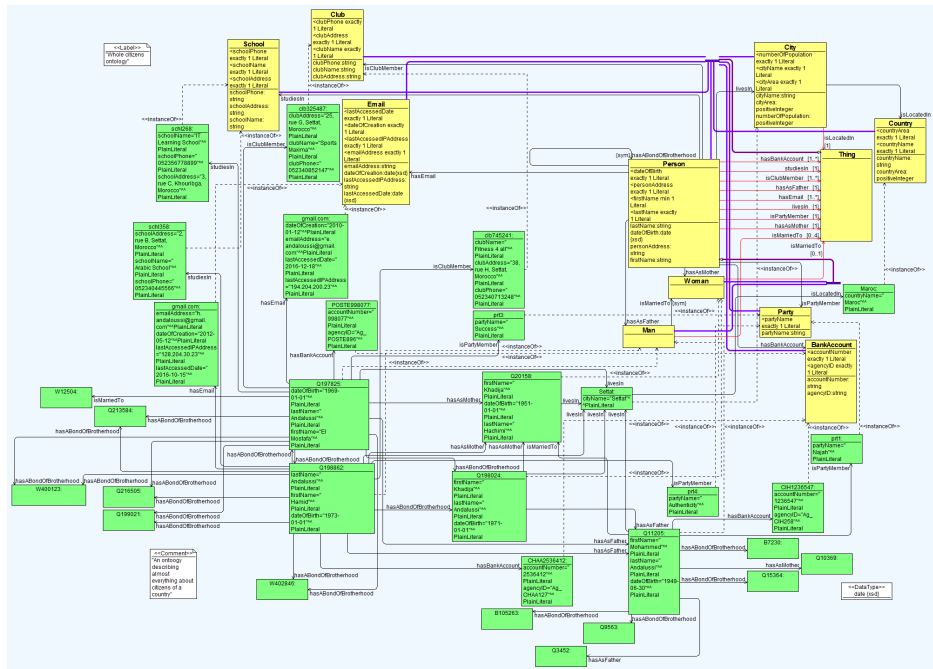

**Fig. 5.** Final Resultant Sub-Ontology Diagram (Classes + Individuals) after purification

## 4     The vertical Segmentation Algorithm

In this type of segmentation, we keep untouched the extension of the otology, i.e. the individuals, and we work on its schema, i.e. the classes and properties. The goal here is to keep a view on parts of the ontology that may concern the end user. For example, we can have a great ontology of citizens that contains almost everything about them. A school may enrich this ontology by information concerning a citizen's studies, i.e. his classes, teachers, classmates, subjects, marks, etc. A cultural club may add some personal information about this citizen, e.g. his hobbies, the books he borrowed, etc. and a bank will be concerned by his financial information.

In spite of duplicate the whole ontology structure in all the sites that will work on it, which leads to more latency when querying such big ontologies with limited resources in local machines, we propose to implement in each site a segment of the ontology that contains only the desired information.

A constraint that should be satisfied when segmenting ontologies vertically is to maintain a link between all the segments, so that we can reassemble them into a whole ontology. There are two kinds of vertical segmentation. One is based on the selection of properties and the other is based on the selection of the classes.

If the segmentation process concerns the properties, then we can use the class ID (<owl:Class rdf:ID="***">) which exists in all the segments to reassemble the whole class definition. On the other hand, if the segmentation process concerns the classes themselves, then we should keep in each segment an object property between a class of this segment and another class in another segment, so that we can reconstruct the whole ontology.

Another kind of vertical segmentation can be that combining both previous kinds, i.e. a segmentation based on both the classes and the properties.

### 4.1 Vertical segmentation based on properties selection

In the vertical segmentation based on the selection of the properties, we apply a filtering process to eliminate some properties that are considered of no interest. For example, and end user may be interested in a citizen's city name but not in its area or its number of population. A club may be interested in a citizen's email address for communication purposes but not in the email's detailed information such as dateOfCreation, lasAccessedDate, lastAccessedIPAddress, etc. For such reasons, we will remove these properties in the extracted sub-ontology.

### 4.2 Vertical segmentation based on classes selection

When the process of segmentation is based on the selection of the classes, we will keep in the sub-ontology only the desired classes. If we will implement the sub-ontologies in different sites with different interests, all the sites will not have the same schema of the ontology. Some of the classes containing the source ontology schema may appear in many sub-ontologies and even in all sub-ontologies, but other classes may be concerned by a unique site. If we think that at some point we will reassemble the different sub-ontologies in order to recreate the global ontology, we must keep links in the sub-ontologies that will allow them to be reassembled. These links are expressed by the objectProperties as in the following case:

```
<owl:ObjectProperty rdf:ID="hasBankAccount">
    <rdfs:domain rdf:resource="#Person" />
    <rdfs:range rdf:resource="#BankAccount" />
</owl:ObjectProperty>
```

which creates a link between the Person and the BankAccount classes.

### 4.3 Vertical segmentation based on both classes and properties selection

Let us continue on the same example of ontology seen in Figure 1. If we want to implement this ontology in different sites with different purposes, then we should extract from this ontology, for each site, a segment containing only the desired information. For example, Figure 6 shows the sub-ontology structure extracted from the citizen's ontology to serve as an ontology for the schools. A school is supposed not to be interested in one's relatives or political tendencies.

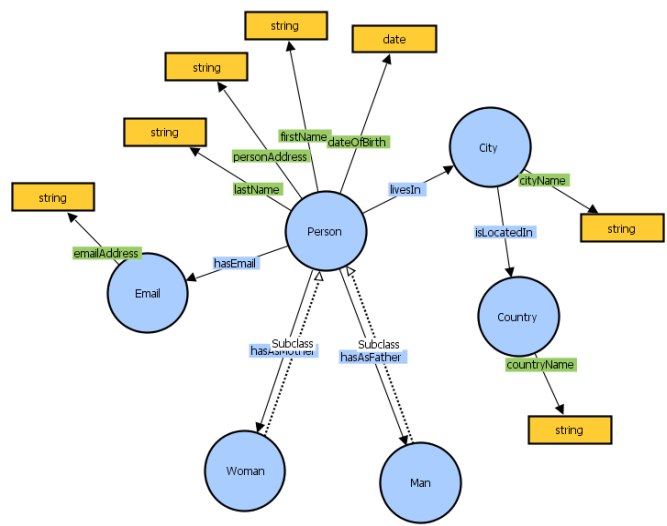

**Fig. 6.** Ontology segment structure extracted from the citizen's ontology to implement in schools

We applied here a double filtering process. First, we removed the classes that will not be in schools' interest, then we removed from the remaining classes all the properties that are considered undesirable for the schools. Tables 1-a and 1-b illustrate classes in Citizen and Sub-Citizen ontologies, where Tables 2-a and 2-b illustrate properties of the Person class in these ontologies.

**Table 1.a** Classes in Citizen ontology.

| Person | Man | Woman | City | Country |
|--------|-----|-------|------|---------|
| Email | BankAccount | School | Club | Party |

**Table 1.b** Classes in Sub-Citizen ontology

| Person | Man | Woman | City | Country | Email |
|--------|-----|-------|------|---------|-------|

**Table 2.a** Properties of the Person Class in Citizen ontology

| hasAsFather | hasAsMother | lastName |
|-------------|-------------|----------|
| firstName | dateOfBirth | livesIn |
| personAddress | hasEmail | hasBankAccount |
| studiesIn | isClubMember | IsPartyMember |

**Table 2.b** Properties of the Person Class in Sub-Citizen ontology

| hasAsFather | hasAsMother | lastName | firstName |
|-------------|-------------|----------|-----------|
| dateOfBirth | livesIn | personAddress | hasEmail |

After extracting this sub-ontology, we may populate it with individuals from the source ontology. This population process should be then purified by removing references to individuals that no more exist. Figures 7 and 8 illustrate the sub-ontology before and after purification process.

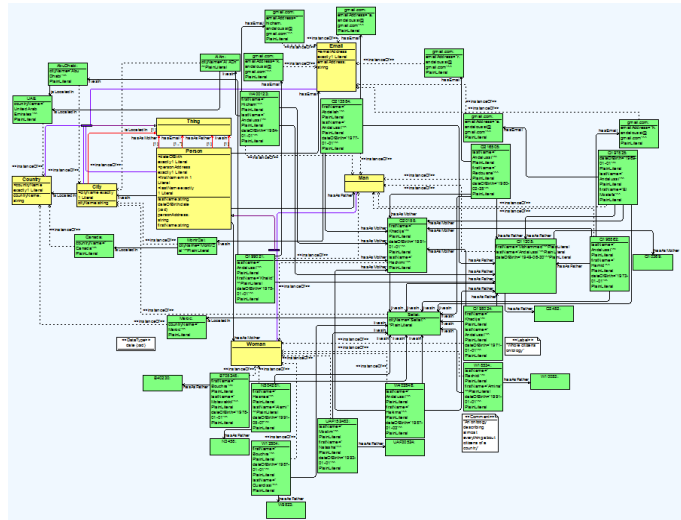

**Fig. 7.** Sub-ontology diagram (classes and individuals) before purification

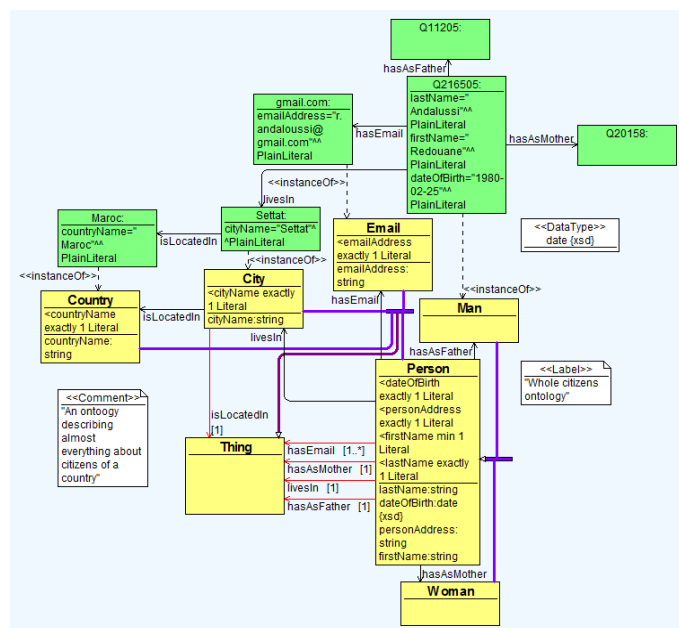

**Fig. 8.** Sub-ontology diagram (classes and individuals) after purification

## 5    Conclusion

The interest of the segmentation of ontologies is to provide a smaller segment which is itself an ontology, so can undergo all treatments of ontologies such as reasoning. The segment has the advantage of the "reduced" size occupied either on disk or on memory, the reduced time necessary for its transfer on the Internet, as well as the reduced time taken by the processor for its treatment.

We presented in this paper different kinds of segmentation. These kinds can be divided into two categories, horizontal and vertical segmentation. The latter can also be considered as two types, based on the properties or the classes. A hybrid vertical segmentation algorithm may combine both methods by filtering the ontology based on its classes and its classes' properties. Also, we may be interested in a full hybrid algorithm where we will combine all the methods by first applying a vertical segmentation based on the classes and the properties selection, then applying a horizontal segmentation based on the properties comparison.